\documentclass[sigconf]{acmart}

\renewcommand\footnotetextcopyrightpermission[1]{} 
\usepackage{algorithm}
\floatname{algorithm}{Algorithm}
\usepackage{algorithmicx} 
\usepackage{algpseudocode}
\newcommand{\methodname}{{\tt{GPS-AFL}}}

\usepackage{xcolor}
\AtBeginDocument{%
  \providecommand\BibTeX{{%
    \normalfont B\kern-0.5em{\scshape i\kern-0.25em b}\kern-0.8em\TeX}}}

\setcopyright{none}
\copyrightyear{2024}
\acmYear{2024}
\acmDOI{XXXXXXX.XXXXXXX}

\acmConference[]{Make sure to enter the correct
  conference title from your rights confirmation email}{
  2024}{Singapore, SG}

\begin{document}
\title{Hire When You Need to: Gradual Participant Recruitment for Auction-based Federated Learning}


\author{Xavier Tan}
\affiliation{%
  \institution{Alibaba-NTU Singapore Joint Research Institute, \\Nanyang Technological University}
  \country{Singapore}
}

\author{Han Yu}
\affiliation{%
  \institution{School of Computer Science and Engineering, \\Nanyang Technological University}
  \country{Singapore}
}


\begin{abstract}
The success of federated Learning (FL) depends on the quantity and quality of the data owners (DOs) as well as their motivation to join FL model training. Reputation-based FL participant selection methods have been proposed. However, they still face the challenges of the cold start problem and potential selection bias towards highly reputable DOs. Such a bias can result in lower reputation DOs being prematurely excluded from future FL training rounds, thereby reducing the diversity of training data and the generalizability of the resulting models. To address these challenges, we propose the \underline{G}radual \underline{P}articipant \underline{S}election scheme for \underline{A}uction-based \underline{F}ederated \underline{L}earning (\methodname{}). 
Unlike existing AFL incentive mechanisms which generally assume that all DOs required for an FL task must be selected in one go, \methodname{} gradually selects the required DOs over multiple rounds of training as more information is revealed through repeated interactions. It is designed to strike a balance between cost saving and performance enhancement, while mitigating the drawbacks of selection bias in reputation-based FL. Extensive experiments based on real-world datasets demonstrate the significant advantages of \methodname{}, which reduces costs by 33.65\% and improves total utility by 2.91\%, on average compared to the best-performing state-of-the-art approach.
\end{abstract}



\maketitle
\pagestyle{plain}

\section{Introduction}
Federated learning (FL) \cite{Tan-et-al:2022TNNLS,Shi-et-al:2023TNNLS,Lyu-et-al:2022TNNLS} is a popular distributed machine learning (ML) paradigm that facilitates collaborative model building involving multiple data owners (DOs), a.k.a. participants, without sharing sensitive local data \cite{lim2020federated}. 
The performance of FL models depends on the quantity and quality of participating DOs \cite{nishio2019client}. It has been recognized that incentives are necessary to motivate participants to contribute to FL consistently in the long term \cite{zhan2021survey}. Given the privacy-preserving nature of FL, it is difficult to directly access the participants' data and resources to determine the optimal amount of incentive payout. As a result, recent FL research works start to leverage DO reputation as a criterion for FL incentivization and participation selection \cite{wang2020novel, tan2022reputation, kang2019incentive, zhang2021incentive, xu2020reputation}.
 
Currently, reputation-based FL faces two key challenges. The first is the cold start problem. Reputation building requires prior interactions between federations and participants. If the selection criteria require DOs to achieve certain minimum reputation scores, it is a challenge for newly joined DOs to have access to FL model training opportunities in order to build their reputation.
Secondly, reputation-based FL participant selection can introduce bias into the resulting models. 
The general preference for highly reputable DOs can introduce biases in the models by repeatedly engaging a small group of them. This can be detrimental to an FL ecosystem due to herding on reputable DOs \cite{yu2013reputation,yu2014reputation,yu2016mitigating,yu2017SciRep}. It can also reduce the diversity of data samples FL models are trained upon, thereby negatively impacting their generalizability \cite{lim2020hierarchical}. Furthermore, existing works do not allow for reputation redemption. Thus, if a DO's reputation falls below a threshold, it is precluded from future opportunities to participate in FL and therefore, unable to build its reputation through self-improvement. Over the long run, these limitations hurt the sustainability of an FL ecosystem.

To address these challenges, we leverage the reverse auction-based FL (AFL) setting which offers a useful way to efficiently reveal DOs' price expectations \cite{ farhadi2023optimal} and promote desirable behaviour. 
Although AFL rewards can motivate DOs to contribute their resources, existing approaches assume that all DOs required for an FL task must be selected in one go. Thus, they employ static participant selection strategies, without cost optimization \cite{xavier2023roboafl, tang2023utility, gupta2023federated}. In view of these limitations, we propose the \underline{G}radual \underline{P}articipant \underline{S}election scheme for \underline{A}uction-based \underline{F}ederated \underline{L}earning (\methodname{}). It gradually selects the required DOs over multiple rounds of training as more information is revealed through repeated interactions. \methodname{} is designed to strike a balance between cost saving and performance enhancement, while mitigating the drawbacks of selection bias in reputation-based FL. 
Through extensive experiments based on 3 real-world datasets, we demonstrate the significant advantages of \methodname{}. Compared to the best-performing state-of-the-art approach, it achieves 33.65\% lower costs and 2.81\% higher total utility on average. To the best of our knowledge, it is the first opportunistic gradual reputation-aware FL participant selection approach designed for AFL settings.


\section{Related works}
\label{related}
Procurement auction, more popularly known as reverse auction, is a process where service providers (i.e., data owners) strive to outperform their competitors by submitting bids with the lowest feasible prices for the goods (i.e., training data) or services to be offered in the auction \cite{farhadi2023optimal}. The reverse auction setting allows for the auctioneer (i.e., FL servers) to achieve desirable outcomes such as budget feasibility, utility maximisation and cost minimization \cite{farhadi2023budget}. The papers \cite{zhang2021incentive, zhang2022online} and \cite{xiong2023truthful} introduced incentive mechanism based on reverse auction and reputation models. The proposed approaches were able to achieve high computation efficiency, and satisfy criteria such as computation efficiency, incentive compatibility (IC), individual rationality (IR) as well as budget constraints. 
In some of these works, DO reputation has been used as a metric to reflect contributions to FL models \cite{wang2020novel, gao2021fifl, tan2022reputation, kang2020reliable}. Authors of \cite{gao2021fifl} also considered DOs' contributions based on their model update gradient so as to allocate rewards fairly. In \cite{kang2020reliable}, DO reputation values are calculated by assessing the frequency of reliable and unreliable interactions with different FL servers. The FL servers verify the reliability and approximate DOs' contribution based on their local computation time and dataset sizes.

These approaches for building reputation requires prior knowledge about and/or interactions with the DOs. 
The approach of recruiting all necessary DOs for an FL task in one go risks over exposure to DOs with unknown reputation in the beginngin stage, which might result in excessive expenditure for relatively low returns. 
Moreover, their methods of participant selection does not allow DOs to redeem their reputation through subsequent self improvement. In addition, these mechanisms are only designed to maximize the procurer's benefit, leading to the depletion of budgets to recruit potentially more than the necessary number of highly reputable DOs. Such a overwhelming preference for highly reputable DOs can also lead to long waiting time for FL tasks due to herding \cite{tan2022reputation}, and potential biases in the resulting models. The proposed \methodname{} approach addresses these limitations.


\begin{figure*}[t]
    \begin{center}
    \resizebox{0.99\textwidth}{!}{\includegraphics{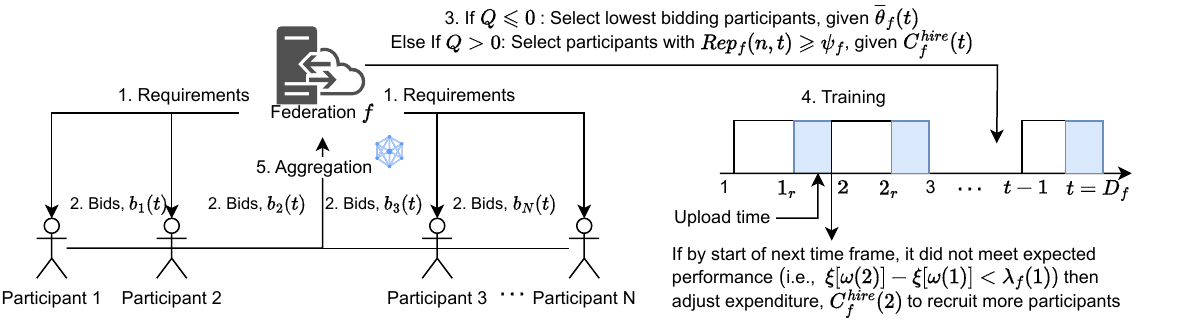}}
    \end{center}
    \caption{Overview of the proposed \methodname{} gradual FL participant selection process.}
      \Description{overview of GPSS}
    \label{fig:overview}
\end{figure*}

 \section {System Model of \methodname{}}
\label{systemmodel}
We consider the setting of a reverse auction-based FL system. The system consists of a finite set of FL servers (i.e., federations) $\mathcal{F} = \{ 1, \ldots , f, \ldots ,F \}$. Each federation is in charge of a singular FL task. For each FL task, there exists a hard deadline for completion, $D_f$. After the deadline $D_f$ has passed, if the FL task is not completed, a federation $f$ receives no utility. There exist a finite set of candidate DOs denoted by $\mathcal{N} = \{ 1, \ldots ,n, \ldots , N\}$. Prospective DOs, regardless of their reputation scores, shall not face complete exclusion from selection consideration. All prospective DOs are free to submit their bids, $b_n(t)$, to join FL training and will be given an opportunity to be considered as long as their bids are low enough.

Due to privacy restrictions, it is difficult for federations to confidently predict the capabilities of the DOs (e.g., their computation/communication resources, amount and quality of local data). That being said, federations can gain knowledge about DOs through repeated interactions during the FL model training. This knowledge can be embodied using reputation scores. To leverage this incremental information revelation process, federations can opt to spread out the participant selection process over a period of time and update their decisions based on subsequent revelation of information (e.g., by observing the DOs' impact on FL model performance). Hence, in this paper, we spread the FL training task over $\mathcal{T}$ time steps $\{1, \ldots, t, \ldots, T\}$, where the end of time step $T$ is aligned with the deadline $D_f$. The overview of this process is illustrated in Figure \ref{fig:overview}. In the following, we explain each step in detail, focusing on one federation, $f$.

\textbf{Step 1: Task Creation and Requirement Announcement.} In this stage, the federation, $f$, will create the FL task and curate the task requirements. The objective of the FL task could involve the minimization of the loss function of the global model concerning a specific test dataset. This loss function can be denoted as $\ell_f(\omega_f(t))$, where $\omega_f(t)$ represents the global model at time step $t$. Thereafter, $f$ will determine the budget for each round, denoted as $\theta_f(t)$. At the beginning, this can be achieved by equally dividing the entire budget, $\theta_f$, over all training rounds (i.e.,  $\theta_f(t) = \frac{\theta_f}{T})$. This will also be referred to as the base per round budget, $\overline{\theta}_f(t)$. Once more information is acquired, the federation can make finer adjustments to the budget allocation.

\textbf{Step 2: Bid Submission.} After announcing the rules and requirements, interested DOs (a.k.a., bidders) can submit their bids, $b_n(t)$ to $f$. A bid vector $\mathcal{B}(t)=\{b_1(t), \ldots, b_n(t), \ldots, b_N(t)\}$ is created from all submitted bids for every time step. In later sections, we will discuss how to ensure the participants only submit truthful bids, and how they will be selected.

\textbf{Step 3: Participant Selection.} To ensure completion of an FL task before the deadline $D_f$ and within the overall budget limit $\theta_f$, $f$ strategically recruits reputable participants with lower bids. Designed based on the Lyapunov optimization frame \cite{yu2016mitigating}, \methodname{} helps a given federation control the cost effectiveness of budget expenditure, while adhering to the budget constraint and FL task requirements. Moreover, federations are not obligated to fully utilize the allocated base budget, $\overline{\theta}_f(t)$, in each time step. Any remaining budget can be rolled over to subsequent time steps to hire more reputable participants opportunistically or achieve cost savings. The final per round budget is calculated as:
\begin{equation}
    \theta_f(T) = \overline{\theta}_f(T) + \sum^{T-1}_{t=1} \left[\theta_f(t) - C^{hire}_f(t)- C^{exe}_f(t)\right]
\label{rolloverbudget}
\end{equation}
where $C^{hire}_f(t)$ represents the amount spent on hiring participants during each time step. $C^{exe}_f(t)$ is the cost to perform FL model training with the hired participants. It is worth mentioning that to maintain budget feasibility, $\theta_f(t) \geqslant 0$. In other words, federation $f$ will cease hiring when there is insufficient budget to continue. To maintain competitiveness, both the bid vector and the winning bid value, $b^{win}(t)$, are announced to all DOs who submitted bids. In the initial time steps when information is scarce, $f$ only focuses on selecting participants with the lowest bids rather than the reputable ones, which usually have higher bids \cite{wang2022dynamic, richardson2019rewarding}. As FL training progresses and information about the DOs are gradually accumulated, $f$ gradually switch to following \methodname{} to algorithmically determine the optimal cost allocation. The essence of \methodname{} is that if the FL model is performing well and on track to be completed by $D_f$, the federation shall minimize spending; otherwise, the federation spend more of the available budget to hire more reputable participants to improve the FL model performance. This strategy is rooted in the notion of opportunistic budget allocation \cite{Yu-et-al:2019AIES,Shi-et-al:2023ICWS}.

\textbf{Step 4: FL Model Training.} The optimization of the global loss function is achieved by minimizing the weighted mean of the participants' local loss functions \cite{mcmahan2017communication}, denoted as $\ell_n(\omega_n(t))$. As such, each selected participant train its local model, $\omega_n(t)$, for a single time step using its own local dataset and compute resources, in attempt to minimize $\ell_n(\omega_n(t))$ within the allocated time. The actual time allocated for training will be shorter than the actual time step $t$ as time has to be allocated for participants to upload their updated model weights. We use $t_{r}$, where $0 < r \leqslant t$, to denote the actual time period used for model training and $t-r$ would be the amount of communication time for participants to upload their model weights.

\textbf{Step 5: FL Model Aggregation.} Participants are required to submit their local model updates to the FL server $f$ at the end of $t_{r}$. $f$ then performs model aggregation following algorithms such as FedAvg \cite{mcmahan2017communication} to update the global model $\omega_f(t)$. Uploads received beyond this time point will not be accepted, and corresponding participants will not receive payment. This is especially important for meeting the deadline $D_f$. The federation then updates participants' reputation scores based on the marginal improvement between global model performance and participant's local model performance. The FL training process continues until either the predefined deadline at $t = D_f$ is reached or until the task is completed, whichever is earlier. 

Thereafter, rewards are distributed to the participants.

During the aggregation phase where FL task has yet been completed, $f$ assesses if the improvement in FL model performance meets or exceeds a predefined performance threshold, $\lambda_f(t)$. If the improvement of model performance, $\xi(\omega_f(t))$, falls below the threshold, $\xi(\omega_f(t))-\xi(\omega_f(t-1))< \lambda(t)$, the federation needs to make adjustments to the hiring expenditure for subsequent time steps. Specifically, the federation shall select more participants with reputation value above a certain threshold, $\psi_f$, to increase the likelihood of attaining the $\lambda_f(t+1)$. If $\xi(\omega_f(t))-\xi(\omega_f(t-1)) \geqslant \lambda_f(t)$, $f$ can reduce hiring expenditure to the minimum in order to save cost. This iterative fine-tuning process of cost expenditure aims to progressively enhance the performance of the FL task, while saving costs.

\subsection{Utility and Cost Model}

\textbf{Utility Maximization}: A federation’s utility, $U_f$, is defined as: 
\begin{equation}
\label{utility}
    U_f(t) = J[\xi(\omega_f(t))] -  C^{exe}_f(t) -  C^{hire}_f(t)
\end{equation}
where $J[\xi(\omega_f(t))]$ represents the revenue yield at time step $t$. The federation can generate revenue using the global FL model by offering services based on the model (e.g., digital banking services \cite{yang2019federated}). Consequently, its utility yield can be represented by the difference between the revenue generated under given global model performance and all costs incurred. $C^{exe}_f(t)$ denotes the execution costs, which consist of communication costs $C^{com}_f(t)$ and computation costs $C^{cmp}_f(t)$. $C^{hire}_f(t)$ represents the total cost of hiring the participants for the entire FL task, which can be derived as:
\begin{equation}
\label{calculatehirecost}
    C^{hire}_f(t) = \sum^N_{n=1} b_n(t) P(n, t), \forall n \in \mathcal{N}
\end{equation}
where $P(n, t) \in [0,1]$ denotes the binary decision vector for selecting the a DO $n$ who submitted bid $b_n(t)$. $P(n, t) = 1$ represents that the DO is selected by $f$ to join FL training at time $t$. Increasing $C^{hire}_f(t)$ implies higher chance of success for completing of FL task before deadline $D_f$, since more participants and possibly more reputable participants can contribute to the FL training. However, based on Eq. \eqref{utility}, it might negatively impact $f$'s overall utility gain. Hence, to maximise $U_f$, the federation has to balance the trade-off between the cost of hiring and the rate of performance improvement. Therefore, we propose a method to strategically determine the value of $C^{hire}_f(t)$ as the sitation facing $f$ changes. More details will be provided in Section \ref{Solution}.
A participant's utility function is defined as:
\begin{equation}
\label{participantutility}
    U_n(t) = [b_n(t) - C_n(t)] \cdot P(n, t) 
\end{equation}
where $C_n(t)$ is its cost of joining FL model training. 

\subsection{Energy Model and Execution Cost Function}
FL model transmission and aggregation incur some computation and communication costs on the FL server. We refer to this as the \textit{execution cost}, which is defined as: 
\begin{equation}
    C^{exe}_f(t) =  \mu_f \cdot E^{sum}_f(t).
\end{equation}
$\mu_f$ is a weight parameter for $f$'s energy consumption. The total energy cost for $f$ in a single time step is: 
\begin{equation}
    E^{sum}_f(t) = E^{com}_f(t)+E^{cmp}_f(t).
\end{equation}
$E^{com}_f(t)$ denotes the energy required by all the selected participants to communicate with federation $f$. $E^{cmp}_f(t)$ is the computation energy required to perform FL model aggregation. In this paper, we adopt the orthogonal frequency division multiple access (OFDMA) protocol \cite{sun2020adaptive} to model the energy consumption of local model updates. $\hat{E}^{com}_f(n, t)$ represents the energy required for participant $n$ to communicate with federation $f$ at time step $t$. 
\begin{equation}
   \hat{E}^{com}_f(n,t) = \frac{\eta_{com} \cdot M} {\sum^U_{u=1} l_{u,n} W \log_2(1+\frac{\rho_{u,n} h_{f,n}}{\iota})}, \forall n \in N.
\end{equation}
$\eta_{com}$ is the normalization parameter for the consumption of communication resources. $M$ denotes the size of the global model $\omega_f(t)$ in bits. $U$ represents the set of uplink sub-channels $\{1 ,\ldots, U\}$. $l_{u,n}$ denotes the time allocated for participant $n$ on sub-channel $u$. As all participants are allocated a fixed time window for uploading their models, $l_{u,n}$ is equivalent to $r$ in the context of local model uploads. $W$ represents the sub-channel bandwidth. $\rho_{u,f}$ denotes the uplink transmission power between $f$ and sub-channel $u$. $h_{f,n}$ denotes the uplink channel power gain between federation $f$ and the participant $n$. Lastly, $\iota$ represents the background noise. The formula for energy consumption to communicate with all participants selected in time $t$ is therefore:
\begin{equation}
    E^{com}_f(t) = \sum^N_{n=1}P(n, t)\hat{E}^{com}_f(n,t), \forall n \in \mathcal{N}.
\end{equation}
The computation energy consumption for the FL server $f$ to perform model aggregation can be expressed as:
\begin{equation}
    E^{cmp}_f(t) = \sum^{N}_{n=1}P(n, t)M_n\varsigma\zeta_fG_f^2 , \forall n \in \mathcal{N}
\end{equation}
where $G_f$ represents $f$'s CPU operating frequency and $\varsigma$ denotes the number of CPU cycles required for FL server to perform the model aggregation for each DO's model update. $M_n$ represents the size of model update provided by DO $n$. Finally, $\zeta_f$ denotes the effective load capacitance of the $f$'s computation chip-set.

\subsection{Reputation Score}
Reputation serves as a useful metric for evaluating participants' data quality, computation and communication capabilities. In this paper, the Beta Reputation System (BRS) \cite{ismail2002beta} is adopted, with binary variables $\alpha^t_n$ and $\beta^t_n$ used to represent the number of positive and negative feedbacks, respectively. A 2-tuple $\langle\alpha^t_n$, $\beta^t_n\rangle$ is used to represent the feedback values from $f$ with regards to the interaction with participant $n$. If $[\xi(\omega_n(t)) \geqslant \xi(\omega_f(t-1))]$ the 2-tuple results to $\langle 1,0\rangle$. Otherwise $[\xi(\omega_n(t)) < \xi(\omega_f(t-1))]$, it would be deemed a negative interaction and tuple updates to $\langle 0,1\rangle$. After participating for $t$ time steps, the feedback values for $n$ updates to $\alpha^t_n = \sum^t_{i=1} \kappa^{t-i} \alpha^i_n$ and $\beta^t_n = \sum^t_{i=1} \kappa^{t-i}\beta^i_n$. The records of reputation scores from previous rounds are denoted as $i$. The parameter $\kappa \in (0,1]$ acts as a discount factor for interaction freshness. Its importance lies in acknowledging that a participant's past reputation might not always reflect their effectiveness in future FL training accurately. 

Therefore, interactions that took place earlier in time between $n$ and $f$ are given reduced significance compared to more recent interactions. When $\kappa$ approaches 1, the weight assigned to interaction freshness remains unchanged, maintaining stronger emphasis on recent interactions. Thereafter the resulting reputation score, $Rep_f(n, t)\in (0,1)$ can then be expressed as:
\begin{equation}
 Rep_f(n, t) = \frac{\sum^t_{i=1} \kappa^{t-i} \alpha^i_n + 1}{\sum^r_{i=1} \kappa^{t-i} \alpha^i_n + \sum^t_{i=1} \kappa^{t-i} \beta^i_n + 2}.
\end{equation}

\section{Optimization Problem and Solution}
\label{Solution}
Based on the discussion above, federations have to balance two main factors when selecting participants: 1) those who bid the lower prices and 2) those who are reputable, in order to ensure that of the FL model can be trained in time, while staying within a given budget.
To achieve these goals, we formulate the optimization problem of \methodname{} as:

Maximize:
\begin{equation}
    \lim_{T\rightarrow\infty}\frac{1}{T} \sum^T_{t=1}  U_f(t)
\label{firstoptimization}
\end{equation}

Subject to:
\begin{equation}
C^{exe}_f(t) +  C^{hire}_f(t) \leqslant \theta_f(t), \forall t\in T.
\label{budgetconstraint1}
\end{equation}

The primary objective is to optimize the average expected utility for $f$ in the long run through a combination of dynamic cost saving and global model performance improvement. Eq. \eqref{budgetconstraint1} ensures that the budget feasibility constraint is fulfilled. For $f$ to complete the FL task before the deadline $D_f$, it becomes imperative to monitor and optimize its cost effectiveness in participant selection as well as the rate of improvement of the global FL model $\omega_f(t)$. We construct a dynamic conceptual queue to capture the regret of $f$. 
The notion of regret involves evaluating how $f$ allocates resources imprudently, resulting in a sub-optimal rate of improvement for the global model with respect to the expected improvement at the present time step. This queue can be expressed as:
\begin{align}
\label{BigQ}
    & Q_f(t+1) = \max[0, Q_f(t) + C^{opt}_f(t)\textbf{I}_{[\xi(\omega_f(t+1)) - \xi(\omega_f(t)) < \lambda_f(t)]}  \nonumber \\
    & - C^{hire}_f(t)], \forall t \in \mathcal{T}
\end{align}
where $C^{opt}_f(t)$ refers to the optimal expenditure for $f$ at time step $t$. The optimal expenditure can be derived by postulating that $f$ only recruits reputable DOs. Reputable DOs can be selected using a reputation threshold value (i.e., only those with reputation values above the threshold value, $\psi_f$). It is worth noting here that it is possible for $C^{opt}_f(t)$ to become larger than $\theta_f(t)$. In such a case, $C^{hire}_f(t)$ is capped by the budget constraint following Eq. \eqref{budgetconstraint1} and Eq. \eqref{budgetconstraint}. 

When the condition $\xi(\omega_f(t+1)) - \xi(\omega_f(t)) < \lambda_f(t)$ is satisfied, the indicator function $\textbf{I}_{[condition]} = 1$. Otherwise, $\textbf{I}_{[condition]} = 0$. $\lambda_f(t)$ is the threshold requirement for model performance improvement. For simplicity of expression, we abbreviate $\textbf{I}_{[condition]}$ to $\textbf{I}$. For all $f\in F$ to complete the FL task before the deadline $D_f$, they must not accumulate too much ``regret''. Thus, we use the $L_2$ norm loss function to measure the amount of queue backlog within the FL system, which can be expressed as:
\begin{equation}
\label{l2norm}
L(t) \triangleq \frac{1}{2}\sum_{f\in F}[Q_f^2(t)]. 
\end{equation}
To establish a bound on the expected increase of $L(t)$ over the long run, we introduce the concept of Lyapunov drift, $\Delta(t)$, which is formulated as: 
\begin{equation}
\label{deltadef}
    \Delta \triangleq \frac{1}{T}\sum^T_{t=1}[ L(t+1) - L(t) ].
\end{equation}
Next, we derive an optimization objective function that combines efforts in increasing utility yield and drift control within a specified bound, resulting in a joint \textbf{\{Utility - Drift\}} objective function as:
\begin{equation}
\label{UtilityminusDrift}
  V \cdot [ {U}_f(t)|C^{exe}_f(t), C^{hire}_f(t)] - \Delta,
\end{equation}
which is to be maximised. The parameter $V > 0$ controls the overall emphasis placed on yielding higher utility over minimising the cost to hire. A higher $V$ value can be understood as stronger emphasis on gaining short term utility through aggressive hiring, possibly to meet tight deadlines; while a lower $V$ value means that $f$ places stronger emphasis on the prudence of budget expenditure. Based on Eq. \eqref{BigQ} and Eq. \eqref{l2norm}, the average Lyapunov drift can be expressed as:
\begin{align}
    & \Delta = \frac{1}{T} \sum^{T}_{t=1}\sum_{f\in F} \biggr( \frac{1}{2}Q^2_f(t+1) - \frac{1}{2}Q^2_f(t) \biggr)  \nonumber \\
    & = \frac{1}{T} \sum^{T}_{t=1}\sum_{f\in F} \biggr( \frac{1}{2} \max \biggr[0, Q_f(t) + C^{opt}_f(t)\textbf{I} - C^{hire}_f(t) \biggr ]^2 \nonumber \\
    &  -\frac{1}{2}Q^2_f(t)  \biggr) \nonumber \\
    & \leqslant \frac{1}{T} \sum^{T}_{t=1}\sum_{f\in F}  \biggr( \frac{1}{2} [Q_f(t) + C^{opt}_f(t)\textbf{I} - C^{hire}_f(t)]^2 -\frac{1}{2}Q^2_f(t) \biggr) \nonumber \\
    & \leqslant \frac{1}{T} \sum^{T}_{t=1}\sum_{f\in F} \biggr(Q_f(t) \biggr[ C^{opt}_f(t)\textbf{I} - C^{hire}_f(t)  \biggr]  \nonumber \\ 
    & + \frac{1}{2}(C^{opt}_f(t))^2\textbf{I} - C^{opt}_f(t)\textbf{I}{C^{hire}_f(t)} + \frac{1}{2}({C^{hire}_f(t)})^2  \biggr) 
    \label{finaldelta}
\end{align}
Throughout the process, we can only directly control the expenditure on hiring, $C^{hire}_f(t)$, to indirectly influence the quantity and quality of participants selected. Since the only variable that affects $C^{exe}_f(t)$ is the number of participants hired, which is of a much smaller scale than $C^{hire}_f(t)$, we only retain variables that contains $C^{hire}_f(t)$. Substituting Eq.  \eqref{finaldelta} with Eq. \eqref{UtilityminusDrift}, we have:

Maximize:
\begin{align}
    & \lim_{T\rightarrow\infty}\frac{1}{T} \sum^{T}_{t=1}\sum_{f\in F}  - VC^{hire}_f(t) - \biggr[-  {C^{hire}_f(t) Q_f(t)}   \nonumber \\ 
    & - {C^{opt}_f(t)C^{hire}_f(t)} + \frac{1}{2}({C^{hire}_f(t)})^2  \biggr]
    \label{finalobjfunc}
\end{align}

Subjected to:
\begin{equation}
    C^{hire}_f(t) + C^{exe}_f(t) \leqslant \theta_f(t)
    \label{budgetconstraint}
\end{equation}

By setting the first-order partial derivative of Eq. \eqref{finalobjfunc} w.r.t $C^{hire}_f$ to be equal to 0, we obtain the upper bound for setting the expenditure when $Q(t) > 0$,
\begin{align}
    & C^{hire}_f(t) =  \left\{ \begin{array}{rcl}
         Q_f(t)+ C^{opt}_f(t)- V & \mbox{,}
         & C^{hire}_f(t) \leqslant \theta_f(t) \\ 
         \theta_f(t) & \mbox{,} & C^{hire}_f(t) > \theta_f(t) \\
        \end{array}\right.
    \label{updatebudget}
\end{align}

Should the evaluated value of $C^{hire}_f(t)$ exceeds the allocated budget $\theta_f(t)$, as determined by Eq. \eqref{rolloverbudget}, the former will be limited to the value of $\theta_f(t)$. This ensures budget feasibility. When $Q(t) \leqslant 0$, the approach will revert to selection based on base budget, $\hat{\theta_f(t)}$, and the computation of $C^{hire}_f(t)$ will follow Eq. \eqref{calculatehirecost}. Algorithm \ref{alg:GPSSAlgo} illustrates how \methodname{} performs FL participant selection. The intuition is that \textit{
The algorithm considers optimal budget expenditure and the rate of improvement for the global model. If $f$ spends more than the optimal expenditure $C^{opt}_f(t)$ while achieving sub-optimal performance gains, the $Q_f(t)$ value increases. The budget allocation is adjusted based on the $Q_f(t)$ value and $C^{opt}_f(t)$ at each time step, within the given budget. The $Q_f(t)$ value represents the difference between the optimal hiring cost and the actual cost. When $Q_f(t) \leqslant 0$, the algorithm prompts $f$ to hire only the lowest bidding participants while not exceeding the base per round budget $\overline{\theta}_f(t)$. It is worth noting that, in reverse auction, the actual hiring cost tends to decrease over time as the lowest bidders are selected. An increasing $Q_f(t)$ value prompts the algorithm to prioritize hiring reputable participants over saving costs. Essentially, \methodname{} opportunistically allocates additional resources when the conditions arise to enhance the overall utility yield.}

\begin{algorithm}[t]
\caption{\methodname{}}\label{alg:GPSSAlgo}
\begin{algorithmic}[1]

\State \textbf{Initialize}: Initial model $\omega_f(0)$, total number of FL training time steps $T$, learning rate $\eta_f$, budget $\overline{\theta}_f(t)$, reputation threshold $\psi_f = 0.7$, and participant bid vector $\mathcal{B}$.

\While{$t < T$}:
    \State $b^{win}(t)$ $\leftarrow$ $\min(\mathcal{B}(t))$;
    \If{$Q_f(t) \leqslant 0$}:
        \ForAll {$b_n (t)$ in $\mathcal{B}(t)$}:
            \If {$b_n(t) == b^{win}(t)$ and $(C^{exe}_f(t) +  C^{hire}_f(t) \leqslant \overline{\theta}_f(t))$}: 
            \vspace{1pt}  Recruit participant $n$;
            \Else: \vspace{1pt}  Reject participant $n$;
            \EndIf
        \EndFor
    \EndIf
    \If{$Q_f(t) > 0$}:
        \State $C^{hire}_f(t) \leftarrow  [{Q_f(t)}+ {C^{opt}_f(t)} - {V}];$
        \If{$C^{hire}(t) > \theta_f(t)$}:
        \State $C^{hire}(t) \leftarrow \theta_f(t)$
        \EndIf
         \ForAll {$b_n(t)$ in $\mathcal{B}(t)$}:
            \If {$(C^{hire}_f(t) - b_n(t) - C^{exe}_f(t) \geqslant 0)$ \& \\ $(Rep_f(n,t) \geqslant \psi_f)$}: 
            \vspace{1pt}  Recruit participant $n$;
            \Else: \vspace{1pt} Reject participant $n$;
            \EndIf
        \EndFor
    \EndIf
\EndWhile
\end{algorithmic}
\end{algorithm}

\section{Analysis}
\label{analysis}

In this section, we perform theoretical analysis on the properties of \methodname{}, truthfulness, individual rationality (IR), computational time complexity, and budget feasibility.

\textbf{Truthfulness}: Participants would have no incentive to report a bid other than their true private valuations under \methodname{}. 
\begin{proof} 
According to Lines 4-9 of Algorithm \ref{alg:GPSSAlgo}, suppose $Q_f(t) \leqslant 0$ and that there is sufficient budget. Assume participant $n$ placed the lowest bid based on its private valuation and wins the round (i.e., $b_n(t)= b^{win}(t)$). If $n$ bids any higher than $b_n$ then $n$ will be rejected resulting in $P(n, t) = 0$ and consequently $0$ utility for that participant. Conversely, if $n$ places a lower bid (i.e., $b'_n(t) < b_n(t)$ where $b'_n(t)$ is the new bid) and wins, then $n$ will receive lower reward than previously possible, resulting in a sub-optimal or negative $U_n(t)$. While $b'_n(t) > b_n(t)$ and $b'_n(t) = b^{win}(t)$ might occur, $n$ would not know $b^{win}(t)$ before the round ends, preventing such bid manipulation. Moreover, due to nature of reverse-AFL, the bid price will eventually decrease. Hence, there is no incentive for any rational participant to not bid truthfully. 
\end{proof}

\textbf{Individual Rationality}: Each participant receives non-negative utility for joining FL model training under \methodname{}. 
\begin{proof} 
When $Q_f(t)\leqslant 0$: if a participant $n$ is selected, Eq. \eqref{participantutility} becomes $U_n(t) = b^{win}(t) - C_n(t)$. Assuming that participant $n$ is rational and bids a value not lower than its cost, $b^{win}(t) = b_n(t)$ and $b_n(t) \geqslant C_n(t)$, then $U_n(t)\geqslant 0$. Otherwise, if $b^{win}(t) \geqslant C_n(t)$ and that $b^{win}(t) \geqslant b_n(t)$, $n$ would not have been selected and will need to wait for the next time step to try again. In this case, $U_n(t) = 0$. In the scenario when $Q_f(t) > 0$, the utility gained by the selected participants would be exactly equal to their bid values. According to Eq. \eqref{participantutility}, since $b_n(t) \geqslant C_n(t)$, $U_n(t) \geqslant 0$.
\end{proof}

\textbf{Computational Time Complexity}: The proposed \methodname{} approach is designed to be lightweight so as not to incur additional burden on the FL system. It achieves a computational time complexity of $O(\mathcal{N})$.
\begin{proof} 
Based on Eq. \eqref{updatebudget} and Algorithm \ref{alg:GPSSAlgo}, the key budget update variable when $Q_f(t)>0$ is $Q_f(t)$, calculated by Eq. \eqref{BigQ} and $C^{opt}_f(t)$. Both of which can be computed with constant time complexity. Hence, based on Algorithm \ref{alg:GPSSAlgo}, \methodname{}'s computational time complexity depends only on the loop to select FL participants from $N$ prospective DOs based on their bids.
\end{proof}

\textbf{Budget Feasibility}: Total cost of hiring the required FL participants under \methodname{} does not exceed the available budget.
\begin{proof} 
According to Lines 6 and 14 of Algorithm \ref{alg:GPSSAlgo} and Eq. \eqref{budgetconstraint}, there is a strict expenditure cutoff threshold for each time slot. As the proposed algorithm adheres to the per-round budget limit for each time slot, it ensures that \methodname{} always remains within the overall budget constraint.
\end{proof}

\section{Experimental Evaluation}
\label{exp}
In this section, we empirically assess the effectiveness of \methodname{} by comparing it to five state-of-the-art approaches using three publicly available datasets. 

\subsection{Experiment Settings}
\subsubsection{Datasets}
The evaluation begins with the popular MNIST dataset \cite{lecun_bottou_bengio_haffner_1998}, containing $70,000$ handwritten digit images, divided into $60,000$ training and the rest for testing. Each image is represented by a $28\times28$-pixel grayscale format. We set our threshold $\lambda_f(t)$ based on unmodified FL training results on MNIST data with the same model parameters. This allows for a comparison of our findings with similar datasets including Fashion-MNIST (FMNIST) \cite{xiao2017fashion} and the EMNIST balanced dataset \cite{cohen2017emnist} using the same threshold values and experiment settings. 
\begin{enumerate}
    \item \textbf{MNIST}: To train on the MNIST dataset, the global model comprises of two convolutional layers using a $5\times5$ kernel size, combined with max-pooling layers of $2\times2$, followed by two fully connected layers with 320 and 50 hidden units, respectively. Stochastic gradient descent with momentum is employed as the optimizer for the training process. 
    \item \textbf{FMNIST}: For the FMNIST dataset, the global model comprises of two similar convolutional layers as before, but an additional batch normalization step is added after each convolutional layer. Followed by three fully connected layers with $1568, 500$ and $200$ hidden units, respectively. The optimization method chosen for both tasks is stochastic gradient descent with momentum of $0.7$, batch size of $128$ and learning rate of $0.01$. The other experimental parameter settings are as shown in Table \ref{tab:parameters}.
    \item \textbf{EMNIST-balanced}: The EMNIST balanced dataset was based on the similarity of its data attributes. It consists of a total of $47$ classes, featuring $112,800$ instances for training and $18,800$ for testing, with balanced distribution of samples for every class. However, due to the difficulty of the task, adjustments were introduced for the training parameters.
\end{enumerate}

\begin{table}[ht]
\caption{GPS-AFL Experiment Settings}
\centering
\begin{tabular}{|r|l|}
\hline
\textbf{Parameters} & \textbf{Value} \\
\hline
Number of participant & $60$\\
Local Epoch(s) & 1 \\
$\theta_f$ & $400$ \\
$\mathcal{T}$ & $80$\\
$t_{r}$ & $1$ \\
$\psi_f$ & $0.7$\\
$V$ & $1$ \\
\hline
\end{tabular}
\label{tab:parameters}
\end{table}

\subsubsection{Comparison Baselines}
Given that our approach operates within an auction framework, we compared our approach against five state-of-the-art approaches that have demonstrated strong performance in the context of AFL, with one that excels in gradual participant selection.
\begin{enumerate}
    \item \textbf{Greedy}: It is a baseline method under the reverse auction setting whereby participants are automatically selected as long as their costs fall below a certain threshold. In our experiments, we set the threshold to be the bid value of the lower $15^{th}$ percentile. This method allows us to evaluate how \methodname{} compares against a straight forward static approach. 
    \item \textbf{Sharpley Value-based Method (SV)} \cite{wang2020principled}: It can be used to evaluate the contribution level of each participant. The method computes a value for each participant's contribution against all other permutations of participant groups. Thereafter, rewards are distributed proportionately based on each participant's associated SV. To handle the computational complexity of calculating SV, an estimation method proposed in \cite{wang2020principled} was implemented. Reputation values are also used to filter the participants in the SV method for a fair comparison with the other approaches. 
    \item \textbf{OORT} \cite{Fan2021oort}: This method gradually and effectively selects participants, while maintaining a balanced approach to address the exploration-exploitation challenge of identifying high-performing participants. we have implemented this method in the AFL context for experimental comparison.
    \item \textbf{ROBO-AFL} \cite{xavier2023roboafl}: It is an algorithm that aims to make optimal trade-off decisions between selecting participants, saving costs and achieving target FL model performance in a dynamic AFL environment. 
    \item \textbf{RRAFL} \cite{zhang2021incentive}: This approach is a reputation-aware incentive mechanism based in reverse auction setting. It selects the maximum number of reputable participants for any given budget constraint.
\end{enumerate} 
There is no budget restriction for `OORT', `SV', `and Greedy' approaches. While, the budget limitation in `ROBO-AFL' relies on the premise that the FL model generates utility at each round, thereafter utilizing the accrued utility from preceding rounds as the budget for subsequent rounds. The `RRAFL' approach similarly features a predetermined budget-based constraint similar to \methodname{}.

\subsubsection{Evaluation Metrics}
We focused on two evaluation metrics in our experiments: 1) \textit{Total Utility}, and 2) \textit{Total Cost}. 
\begin{enumerate}
    \item The \textbf{Total Utility} is calculated as the summation of utilities generated by the model from $T=1$ to $T=80$ ( defined in Eq. \eqref{utility}). To streamline the experiment, we standardized the revenue yield for all methodologies to be equal to the accuracy of the global model's performance. The higher the total utility, the better the performance of a given approach. 
    \item The \textbf{Total Cost} is defined as the cumulative expenditure incurred by federation $f$, encompassing both $C^{exe}_f(t)$ and $C^{hire}_f(t)$, over all experimental time steps. To ensure fair comparison, all participants incur the same amount of $C^{exe}_f(t)$. Therefore, the only variable that differs is $C^{hire}_f(t) > 0$, which is determined by the auction process. The lower the total cost, the better the performance of a given approach.
\end{enumerate} 

\subsection{Results and Discussion}
In this section, we evaluate the performance of \methodname{} and compare it with four other state-of-the-art approaches based on the aforementioned evaluation metrics.

\begin{figure}[!t]
\centering
\includegraphics[width=0.90\linewidth]{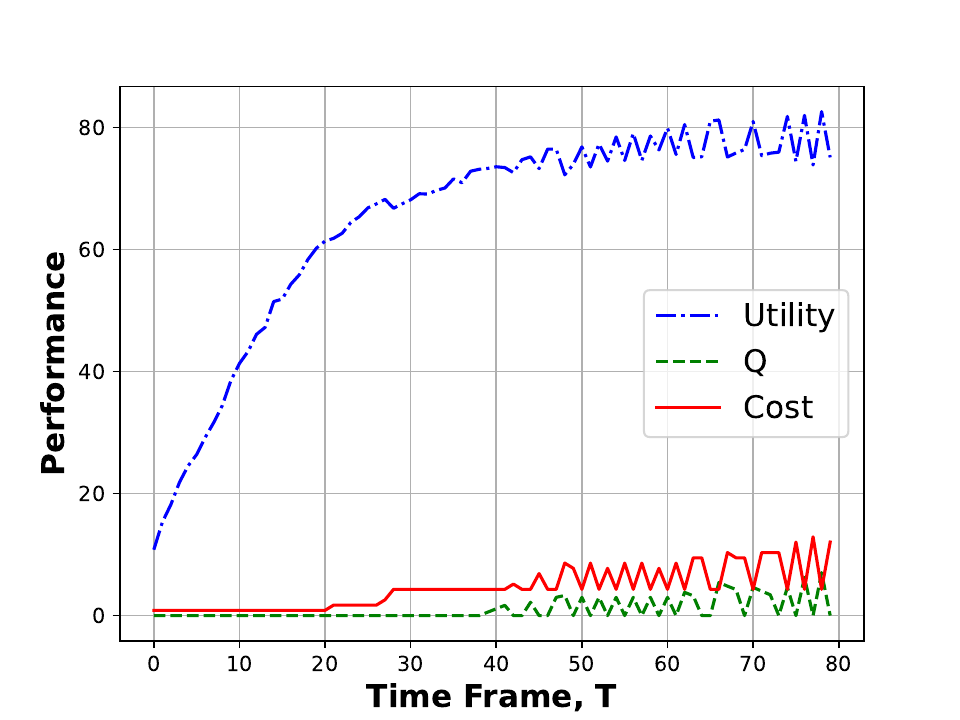}\par
\caption{Utility vs. Cost.}
\Description{}
\label{fig:q_results}
\end{figure}
\begin{figure}[!t]
\centering
\includegraphics[width=0.91\linewidth]{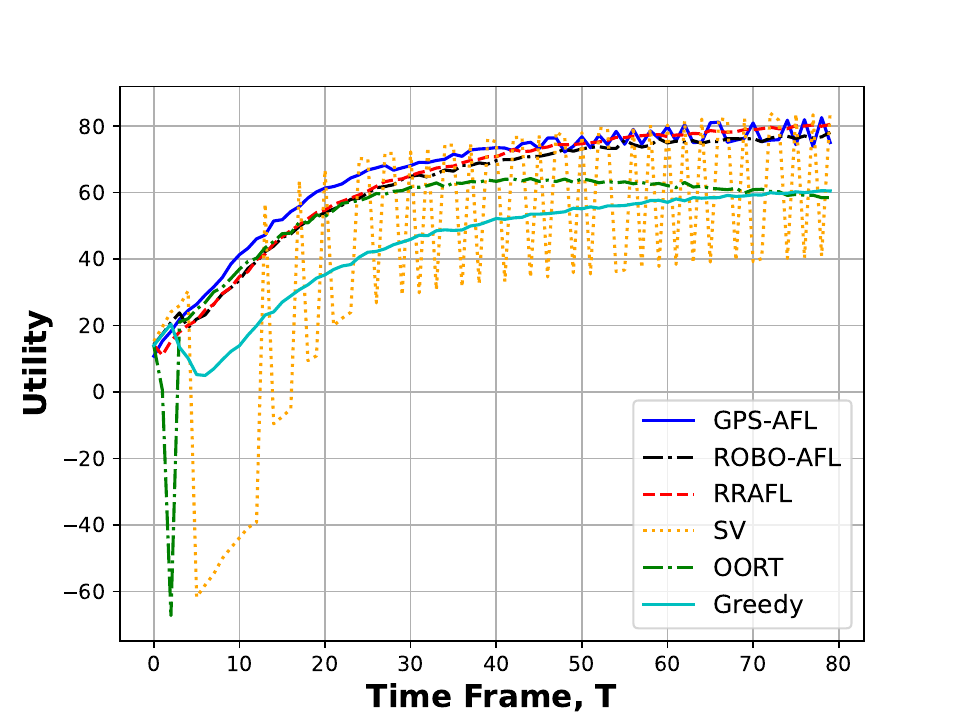}\par
\caption{Comparison of total utilities.}
\Description{}
\label{fig:comparison_results}
\end{figure}
Figure \ref{fig:q_results} depicts the utility yield of federation $f$ in relation to the cost incurred for executing and hiring participants for the completion of the FL task. It can be observed that when the cost spikes up, $Q$ will increase. By virtue of the Lyapunov optimization, \methodname{} will be forced to keep $Q$ low and hence, reduce overall cost. As an added bonus, this figure also reveals that as the model approaches convergence, becoming more sensitive to low-quality data, \methodname{} strategically favours reputable participants (since according to Algorithm \ref{alg:GPSSAlgo}, when $Q>0$ only reputable participants are selected) to reduce the potential harm to FL models. 
\begin{table}[b]
\caption{Comparison results under the MNIST dataset.}
\centering
\begin{tabular}{c|rrr}
\toprule
\textbf{Method} &  \textbf{Total Utility} & \textbf{Total Cost}  \\\hline
Greedy & 3,538.02 & 1,959.65 \\
SV & 3,498.57 & 2,048.84 \\
OORT & 4,277.52 & 1,215.29 \\
ROBO-AFL  & 4,906.35  & 620.11 \\
RRAFL & 5,012.20 & 396.91\\\hline
\methodname{} & \textbf{5,179.71} & \textbf{358.87}\\
\bottomrule
\end{tabular}
\label{tab:exp1}
\end{table}

\begin{table}[t]
\caption{Comparison results under the Fashion-MNIST dataset.}
\centering
\begin{tabular}{c|rrr}
\toprule
\textbf{Method} &  \textbf{Total Utility} & \textbf{Total Cost} \\ \hline
Greedy & 4,430.47 & 2,029.87\\
SV & 5,059.74 & 1,453.61\\
OORT & 5,186.23 & 1,255.67\\
ROBO-AFL & 5,744.60  & 763.47\\
RRAFL & 5,969.95 & 386.99 \\\hline
\methodname{} & \textbf{6,026.60}& \textbf{290.60}\\
\bottomrule
\end{tabular}
\label{tab:exp2}
\end{table}

\begin{table}[t]
\caption{Comparison results under the EMNIST balanced dataset.}
\centering
\begin{tabular}{c|rrr}
\toprule
\textbf{Method} &  \textbf{Total Utility} & \textbf{Total Cost} \\\hline
Greedy & 1,953.05 & 1,192.74 \\
SV & 1,714.22 & 1,078.06\\
OORT & 1,605.30 & 981.76\\
ROBO-AFL & 2,150.37 & 462.63\\
RRAFL & 2,679.22 & 375.46 \\\hline
\methodname{} & \textbf{2,798.20}& \textbf{238.88}\\
\bottomrule
\end{tabular}
\label{tab:exp3}
\end{table}

By examining Figure \ref{fig:comparison_results}, it becomes apparent that \methodname{} consistently attained comparatively superior utility scores in relation to alternative approaches. By about time step $35$, \methodname{} succeeded in attaining over $70$ utility points, while RRAFL managed to reach a comparable value only by approximately time step $45$. Notably, the graphical representation illustrates fluctuations during the concluding rounds, wherein \methodname{} is engaged in the delicate act of balancing between cost and performance optimization. The superiority of this phenomenon becomes evident when comparing the total utility and cost, as shown in the tables below. From Tables \ref{tab:exp1} and \ref{tab:exp2}, we can observe that our algorithm, \methodname{}, performs very well when compared to other state-of-the-art approaches. 

Table \ref{tab:exp1} depicts the simulation results of the approaches under the MNIST dataset.  \methodname{}'s scored a total utility of $3.23\%$,  $5.57\%$, $21.10\%$ ,$46.40\%$, $48.05\%$ higher than `RRAFL', `ROBO-AFL', `OORT', `Greedy' and `SV' approach respectively. While simultaneously incurring $10.60\%$, $72.80\%$, $238.64\%$, $446.06\% $ and $470.91\%$ lesser costs than the other approaches in the same order, respectively. The result demonstrates the adeptness of our algorithm in effectively maximizing budget utilization while achieving an improved balance between cost and performance. As `RRAFL' opts for the largest feasible participant cohort within a stipulated budget for each round, there inevitably arises extra budget unnecessarily spent. This extra expenditure even when the model is performing well constitutes to cost wastage, as opposed to \methodname{} where such residual unspent budget is carried forward to the subsequent round. Furthermore, `OORT' was designed to identify and select high-performing participants without any budget consideration. This consequently led to increased overall costs despite initial utility improvements, as depicted in Figure \ref{fig:comparison_results}.

Table \ref{tab:exp2} depicts the simulation results of the approaches under the FMNIST dataset, \methodname{} also outperformed other approaches. It was able to obtain a total utility of $56.64$,  $282$, $840.37$, $966.87$ and $1596.13$ more points higher than `RRAFL', `ROBO-AFL', `OORT', `SV' and `Greedy' approaches, respectively. While simultaneously incurring $33.17\%$,  $162.70\%$, $332.10\%$, $400.21\% $ and $598.51\%$ lesser costs than the other approaches in the same order, respectively. Likewise, \methodname{} effectively optimized its utility score output to a even higher degree. This could plausibly be attributed to the larger FL model used, which enhanced learning and performance capabilities.

Table \ref{tab:exp3} presents the experimental result concerning a more difficult FL task, based on the EMNIST balanced dataset. The number of participants available has been reduced to $20$. To save on communication costs, each participant will train for $3$ local epochs, with batch size $64$. Consequently, $t_{r}$ tripled to become $3$. The remaining parameters remained unaltered. The FL model deployed for this FL task is identical to the one used in the FMNIST FL task. It is worth noting that our approach can be extended to other FL tasks by leveraging on alternative available benchmarks like the open-sourced benchmarking framework offered by \cite{zeng2023fedlab}.  As shown in Table \ref{tab:exp3}, it remains evident that \methodname{} maintains its superiority over other state-of-the-art methodologies while keeping costs low. \methodname{} was able to attain a total utility of $4.44\%$,  $30.12\%$, $74.31\%$, $63.23\%$ and $43.27\%$ more points higher than `RRAFL', `ROBO-AFL', `OORT', `SV' and `Greedy'  approaches, respectively. While simultaneously incurring $36.38\%$, $57.18\%$, $310.98\%$, $399.31\%$ and $351.30\%$ lesser costs than the other approaches in the same order, respectively.

\section{Conclusions and Future Work}
\label{conclude}

The paper introduces \methodname{}, a method designed to optimize limited budget resources efficiently for constructing high-performance FL models in dynamic reverse-AFL markets. \methodname{} enables federations to select reputable participants as needed and permits them to redeem their reputation when conditions are favourable. Theoretical analysis and experimental evaluation both demonstrate that \methodname{} exhibits favorable characteristics and outperforms existing approaches by striking an advantageous balance between cost saving and performance enhancement. To the best of our knowledge, it is the first opportunistic gradual reputation-aware FL participant selection approach designed for AFL settings

In this study, we operated under the assumption that there is an absence of competition among federations, thereby eliminating external factors that could influence participants' perception of their costs. In subsequent research, we plan to investigate how to eliminate this assumption and introduce the idea of competition among federations into \methodname{}.

\section*{Acknowledgements}
This research is supported by National Research Foundation, Singapore and DSO National Laboratories under the AI Singapore Programme (AISG Award No: AISG2-RP-2020-019); Alibaba Group through Alibaba Innovative Research (AIR) Program and Alibaba-NTU Singapore Joint Research Institute (JRI), Nanyang Technological University, Singapore; and the RIE 2020 Advanced Manufacturing and Engineering (AME) Programmatic Fund (No. A20G8b0102), Singapore.






\bibliographystyle{ACM-Reference-Format}
\bibliography{main}










\end{document}